\documentclass{article}

\usepackage{arxiv}

\usepackage[utf8]{inputenc} 
\usepackage[T1]{fontenc}    
\usepackage{hyperref}       
\usepackage{url}            
\usepackage{booktabs}       
\usepackage{amsfonts}       
\usepackage{nicefrac}       
\usepackage{microtype}      
\usepackage{lipsum}
\usepackage{graphicx}
\usepackage{amsmath}
\usepackage{float}
\usepackage{subfig}
\usepackage[ruled,vlined,noend]{algorithm2e}
\usepackage[noend]{algpseudocode}
\graphicspath{ {./images/} }

\title{FedFMC: Sequential Efficient Federated Learning on Non-iid Data}

\author{
 Kavya Kopparapu \\
  Department of Computer Science\\
  Harvard University\\
  Cambridge, MA 02138 \\
  \texttt{kavyakopparapu@college.harvard.edu} \\
   \And
 Eric Lin \\
  Department of Computer Science\\
  Harvard University\\
  Cambridge, MA 02138 \\
  \texttt{eric\_lin@college.harvard.edu} \\
}

\begin{document}
\maketitle
\begin{abstract}
    As a mechanism for devices to update a global model without sharing data, federated learning bridges the tension between the need for data and respect for privacy. However, classic FL methods like Federated Averaging struggle with non-iid data, a prevalent situation in the real world. Previous solutions are sub-optimal as they either employ a small shared global subset of data or greater number of models with increased communication costs. We propose FedFMC (Fork-Merge-Consolidate), a method that dynamically forks devices into updating different global models then merges and consolidates separate models into one. We first show the soundness of FedFMC on simple datasets, then run several experiments comparing against baseline approaches. These experiments show that FedFMC substantially improves upon earlier approaches to non-iid data in the federated learning context without using a globally shared subset of data nor increase communication costs.
\end{abstract}


\section{Introduction}
Over the past decade, improvements in machine learning accuracy has met rising consumer demand for use of learning models. Models were shown to be successful in a plethora of learning tasks, ranging from prediction-based recommendation to a variety of image processing tasks. Standard machine learning applications flourished in optimized processes where data would flow from user devices to centralized data centers for analysis. 

However, the rise of mobile and IoT (internet-of-things) devices has come with a changing dynamic resulting in new needs and constraints for machine learning. These include: (1) tighter restrictions \cite{KrzysztofekMariusz2019G:GD} on usage of data in relation to user privacy, (2) demand for lower latency in many applications such as remote surgery, and (3) constraints of poor internet connectivity for accessibility in under-developed countries and offsite rigs (for research or oil exploration). 

A wide variety of approaches have strived to address these issues. Most early work focused on privacy conscious learning, some of which formed the foundational work of using a centralized server to learn local work without sacrificing privacy \cite{vaidya2008privacy}. In 2016, Federated Learning (FL) was introcued by McMahan et al. \cite{fedavg} to address these needs. It outlined an approach of keeping data distributed on edge devices, running a global model on each device, and incorporating model updates from each device through a novel method named Federated Averaging. Federated Learning has proven to be groundbreaking in many applications, leading to the birth of many avenues for further research.

Although immediately useful in many contexts, Federated Learning and in particular Federated Averaging depend on an iid global dataset for success in learning. Indeed, McMahan et al. cited non-iid data as a major obstacle in their seminal paper \cite{fedavg}. Historically, non-iid data has not been a problem. In normal machine learning workflows, data centers have access to the entire global dataset and thus distributes data among machines in a carefully controlled iid manner.

Unfortunately, the central server in Federated Learning does not have the same luxury as normal data centers as it lacks direct access to data. Since there are many real-world scenarios which local edge devices will not have data drawn IID from some shared global distribution, researchers have struggled to address this problem. Current approaches either make sacrifices by requiring a globally shared subset of data or increasing communication costs with more models.

In this paper, we tackle supervised learning tasks drawing examples $(x,y) \sim \mathcal{P}_i(x,y)$ from non-identical edge device distributions. More specifically, our method handles bias in label skew, when the marginal distributions of labels $\mathcal{P}_i(y)$ varies across edge devices while $\mathcal{P}(y|x)$ is identical \cite{flreview}.

We restrict our definition of non-iid to define edge devices belonging to certain "archetypes," defined by a set of labels of data that is over-represented by a quantity represented by a bias parameter. Therefore, these archetypes represent label skew. 

Our contributions are as follows:
\begin{enumerate}
    \item Present FedFMC (Fork-Merge-Consolidate) as a novel approach for learning non-iid data in FL.
    \item Address shortcomings of previous approaches by not requiring a globally shared subset of data and showing faster convergence with no increasing communication costs compared to baseline methods.
    \item Show soundness of FedFMC in learning data from different sets of archetypes.
    \item Analyze increase in communication costs and model size.
\end{enumerate}

\section{Previous Work}
\label{sec:headings}
\subsection{Federated Averaging (FedAvg)}
Introduced in the seminal paper by McMahon et al., the Federated Averaging algorithm (FedAvg) is the basis of decentralized learning where several edge devices send updates of locally-trained models to a global server. Each device runs a local copy of the global model on its local data. The global model's weights are then updated with an average of device updates and deployed back to the edge \cite{fedavg}. Algorithm 1 details the algorithm of FederatedAveraging, where $K$ is the number of total devices.

\begin{algorithm}
    \SetAlgoNoLine
    \DontPrintSemicolon 
    initialize $w_0$\;
    \For{each round $t = 1,2,\dots$}{
        $S_t \gets$ (random set $S$ of devices)\;
        \For{each client $k \in S_t$}{
            $w_{t+1}^k \gets$ ModelUpdate($k$, $w_t$)\;
        }
        $w_{t+1} \gets \sum_{k=1}^K \frac{n_k}{n} w_{t+1}^k$\;
    }
    \caption{FederatedAveraging}
    \label{algo:max}
\end{algorithm}%

This builds off of previous distributed learning work by not only supplying local models but also performing training locally on each device. Hence, FedAvg empowers edge devices to collaboratively learn a shared prediction model while keeping all training data local.

Although successful in classic Federated Learning tasks, FedAvg falls short in many real-world scenarios. On non-iid data, the FedAvg algorithm is prone to conflicting updates and performance oscillations as the averaged model is likely to prefer a certain archetype over another. Updates from devices of one archetype tend to undo the updates from another, and as such the averaging function in FedAvg struggles to learn a significant portion of the labels. As a result, FedAvg's slow convergence and low accuracy deem it unusable in most non-iid contents.

\subsection{Current Approach to non-iid FL}
A simple and successful approach to non-iid federated learning is to create a globally-shared dataset of separate origin from the edge devices \cite{flreview}. This dataset is globally-shared between all devices. As such, during the FedAvg process devices are able to update to the global model with less conflicts, enabling the network to learn non-iid data. In particular, Zhao et al. showed in \cite{globaldata} that globally sharing 5\% of data resulted in a 30\% increase in accuracy on experiments with the CIFAR-10 dataset. A version of this globally-shared dataset is found in most state-of-the-art solutions to non-iid FL.

Unfortunately, this approach breaches scenarios of strict Federated Learning. It assumes the accessibility and validity of a such subset of data to generalize to the problem at hand. This is inapplicable to many use cases as there are many legal, social, and technical barriers in sharing data representative of all devices. Thus, this approach  may in fact increase the likelihood of unfair device bias and raise susceptibility to attacks through data poisoning.

\subsection{Lifelong Learning and FL}
Another approach is to consider learning a classifier on each edge device as a separate learning problem and apply lifelong learning techniques \cite{flreview}. Lifelong learning is the task of learning separate tasks $f_1, f_2, ... f_n$ sequentially using a single model, without forgetting the previously learned tasks. 

While there have been many recent developments in lifelong learning, a particularly successful one is Elastic Weight Consolidation (EWC). First introduced by Kirkpatrick et. al, EWC is a method to aid the sequential learning of separate tasks using the same model without "forgetting" tasks that were previously learned \cite{forgetting}. EWC identifies the parameters that are most informative for learning a task and penalizes changing these parameters during the training of future tasks \cite{forgetting}. Specifically, for the lifelong learning of two tasks $A$ and $B$, the objective of training task $B$ would be 
\begin{equation}
    \mathcal{L}(\theta) = \mathcal{L}_B(\theta) + \lambda \sum_{i}^{}(\theta-\theta_A^\ast)^Tdiag(\mathcal{I}(\theta_A^{\ast}))(\theta-\theta_A^\ast)
\end{equation}
where $\mathcal{I}(\theta_A^{\ast})$ is the diagonal of the Fisher information matrix \cite{forgetting, edgify}.

The current approaches to applying lifelong learning techniques require that every device be present in every round of training, and full device participation is not necessarily guaranteed in the real world \cite{flreview, edgify}. In addition, these multi-task learning approaches typically involve decentralized or peer-to-peer learning schemes, which also assumes complete edge device- edge device connectivity, which is not possible in most privacy-conscious applications \cite{flreview, peertopeerFL}. 

\section{The Federated Fork-Merge-Consolidate (FMC) Algorithm}
In the line of decentralized peer-to-peer FL, we model the non-iid learning problem as a lifelong learning task. However, instead of considering each edge device as a separate "task" to learn, we design a scheme, FedFMC, to first group devices with similar archetypes together and treat each group as a task to learn. 

The FedFMC algorithm has two phases: forking and merging. Here, we denote $a_n^{(t)}$ and $l_n^{(t)}$ to be the validation accuracy and validation loss, respectively, of device $n$ at time step $t$. $g_n^{(t)}$ is the group that device $n$ belongs to at time step $t$ and $\theta_j^{(t)}$ is the averaged model associated with the group $j$. $h_f$ and $h_m$ are tune-able hyper-parameters for the forking and merging thresholds, respectively. $T$ is the total number of rounds of forking, $K$ is the number of devices participating in each round of a total of $N$ edge devices, and $M$ is the maximum number of rounds per merging of two models. 

Algorithm 2 details our methodology for dynamically forking a single global model for $T$ rounds. Building off of FederatedAveraging, we first train a subset of $K$ devices every round, update the global model accordingly, and send updates to each edge device. Then, in certain rounds $t$ eligible for forking, each device dynamically sorts itself into the best model through a multi-part process. (1) It first compares its validation loss to other devices reporting to the same model. (2) If its loss is notably higher, it runs all other global models to try to find another model that it has a lower loss. (3) If all the other models perform even worse than the model it is using currently, it forks into a new model. This process is done for each device in the network. 

Since devices of similar data skew (archetype and bias) will behave similarly in terms of validation loss, the {\sc Fork} algorithm encourages dynamic grouping of devices. As such, at the end of $T$ rounds of forking, the global network will have devices that self-forked themselves into groups of similar archetypes. Since {\sc Fork} does increase communication costs, a certain number of rounds have to pass between rounds eligible for forking. We analyze this added communication costs in more detail in section 4.5. 

\begin{algorithm}[H]
\SetAlgoNoLine
\SetKwFunction{algo}{algo}\SetKwFunction{proc}{proc}
$\text{Initialize all }g_n^{(1)}\text{ to 1}$\\
    \For{$t = 1, 2, \dots,T$}{
            $\textit{round\_devices} \gets \text{ a random subset of  }K\text{ devices}$\\
            $\text{Train }\textit{round\_devices}\text{ for E local epochs each}$\\
            \For{$f = 1,2, \dots, \text{\upshape number of groups} - 1$}{
                $\textit{w\_avg} \gets \text{AverageWeights}(n \text{ s.t. } g_n^{(T)} == f)$\\
                $\text{All devices } n \in g \gets \textit{w\_avg}$\\
                $\text{Evaluate device } n \text{ on local validation data}$\\
            }
            \If{$t \textbf{ }\text{\upshape  is eligible for forking}$}{
                \For{$n = 1, 2, \dots N$}{
                    \If{$\text{\upshape group}\text{ } g_n^{(t)}\text{\upshape 's }\text{ }\text{\upshape model performs poorly on device}\text{ }n\text{ }\text{\upshape's local dataset}$}{
                        $g_n^{(t+1)} \gets \text{group } j\text{ for which }\theta_j^{(t)} \text{ has the smallest loss on } n \text{'s local validation data}$\\
                        \If{$j = g_n^{(t)}$}{
                            $n \text{ is placed in a new group}$\\
                        }
                    }
                }
            }
    }
\caption{{\sc Fork} the single global model into multiple groups dynamically.}
\end{algorithm}

Algorithm 3 dictates our approach for merging and consolidating several forked models into one. Note that after {\sc Fork}, there are several models that each have learned a separate data skew / archetype. Then, our consolidation process starts off with one model and iteratively merges models into the globally shared model. As mentioned above, we employ the use of Elastic Weight Consolidation to remember previously learned archetypes as we incorporate more archetypes. At the end, we will have one globally shared model that learns all archetypes.

Note that our work in merging and consolidate models into one is well motivated. Past work have sometimes utilized a process of giving every device a separate model and then boosting device accuracies. However, if the models are not consolidated into one, many advantages of Federated learning are lost. For instance, the ability to receive globally shared updates prove useful in scenarios where the edge device can benefit from other data sources. Indeed, without consolidation, there is little difference between those scenarios and devices which train a model locally offline.

\begin{algorithm}
    \SetAlgoNoLine
    \DontPrintSemicolon 
    \KwIn{$T, K, N,$ and $M$ as defined above.}
    \KwOut{$\theta_{\text{consolidated}}$, a single consolidated global model that performs well for data from all devices and their archetypes.}
    \textit{active\_devices} $\gets$ ($n \text{ with } g^{(T)} == 0$)\;
    
    \For{$f = 1,2, \dots$, \text{number of groups} - 1}{
        $i \gets 0$\;
        $\textit{active\_devices} \gets \textit{active\_devices} + (n \text{ s.t. } g_n^{(T)} == f)$\;
        
        $\theta_{\text{consolidated}} \gets \theta_{\text{active\_devices}}$\;
        EWCFisher matrix $\gets \theta_{\text{prev}}$\;
        \While{$i < M$ and $\text{\upshape not  converged}$}{
            $i \gets i + 1$\;
            \For{all \textit{active\_devices}}{
                $w\_avg \gets$ AverageWeights($n \in$ \textit{active\_devices})\;
                All devices $n \in$ \textit{active\_devices} $\gets w\_avg$\;
            }
            $\textit{round\_devices} \gets$ a random subset of \textit{active\_devices}\;
            EWCTrain \textit{round\_devices} for $E$ local epochs each\;
        }
    }
    \Return{$\theta_{\text{consolidated}}$}\;
    \caption{{\sc Merge} and {\sc Consolidate} forked groups into one global model.}
    \label{algo:max}
\end{algorithm}%

\section{Results}
\subsection{Experimental Setup}
We used subsets of the CIFAR-10 Dataset \cite{cifar10data} with a random group of $M = N/2$ devices participating in each round of training. The only comparative baseline with similar communication costs and underlying assumptions is the FedAvg algorithm. For the EWC of the Merging phases, we used a $\lambda = \frac{1}{\text{Number of Groups}}$. 

This experiment was our proof of concept to show the necessity of different parts of the FedFMC algorithm. We reduced the CIFAR-10 dataset to a very simple example: three archetypes each with only one label (0, 1, or 2). In this case, the forking method consistently forks correctly, so the fork groups are equivalent to groups of devices with the same archetype. 

\subsection{Simple Archetypes: 3 Labels}

\begin{figure}[H]
    \centering
    \includegraphics[scale=0.7]{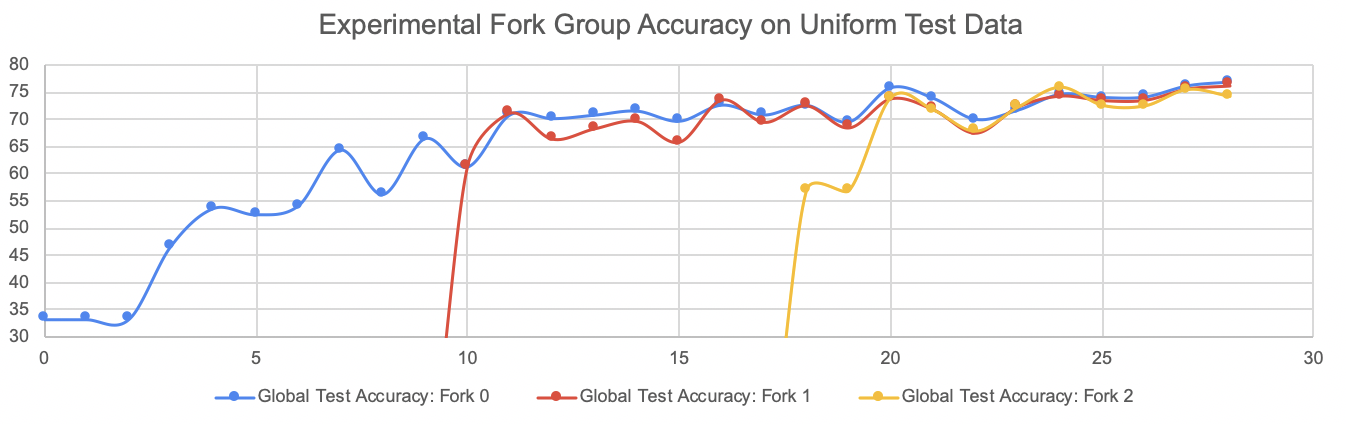}
    \caption{The effects of incorporating EWC from the merging phase of the 3 Archetypes experiment.}
    \label{fig:ewc}
\end{figure}

\begin{figure}[H]
    \centering
    \includegraphics[scale=0.7]{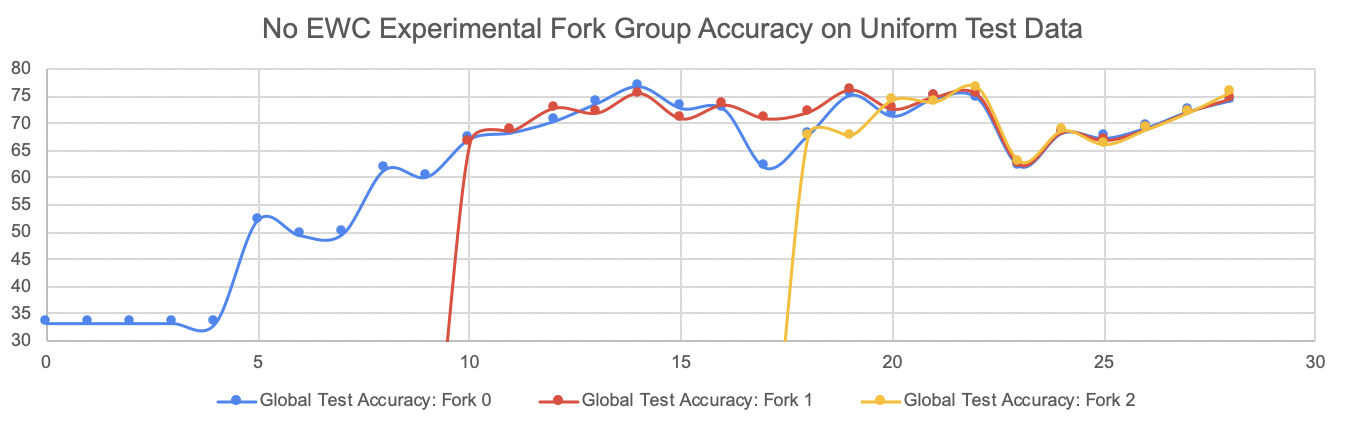}
    \caption{The effects of removing EWC from the merging phase of the 3 Archetypes experiment.}
    \label{fig:ewc2}
\end{figure}

Figures \ref{fig:ewc} and \ref{fig:ewc2} depict the test accuracies of each fork group using the single consolidated global model on a dataset that has uniform representation of all three labels.  It is clear that there is a stabilizing effect of the EWC: without EWC the per-group performance is not stable while learning other groups. As the merging rounds successively incorporated the different groups into a single global model, their performance varied together when the training incorporated EWC (rather than decreasing or oscillating). 

\begin{figure}[H]
    \centering
    \includegraphics[scale=0.47]{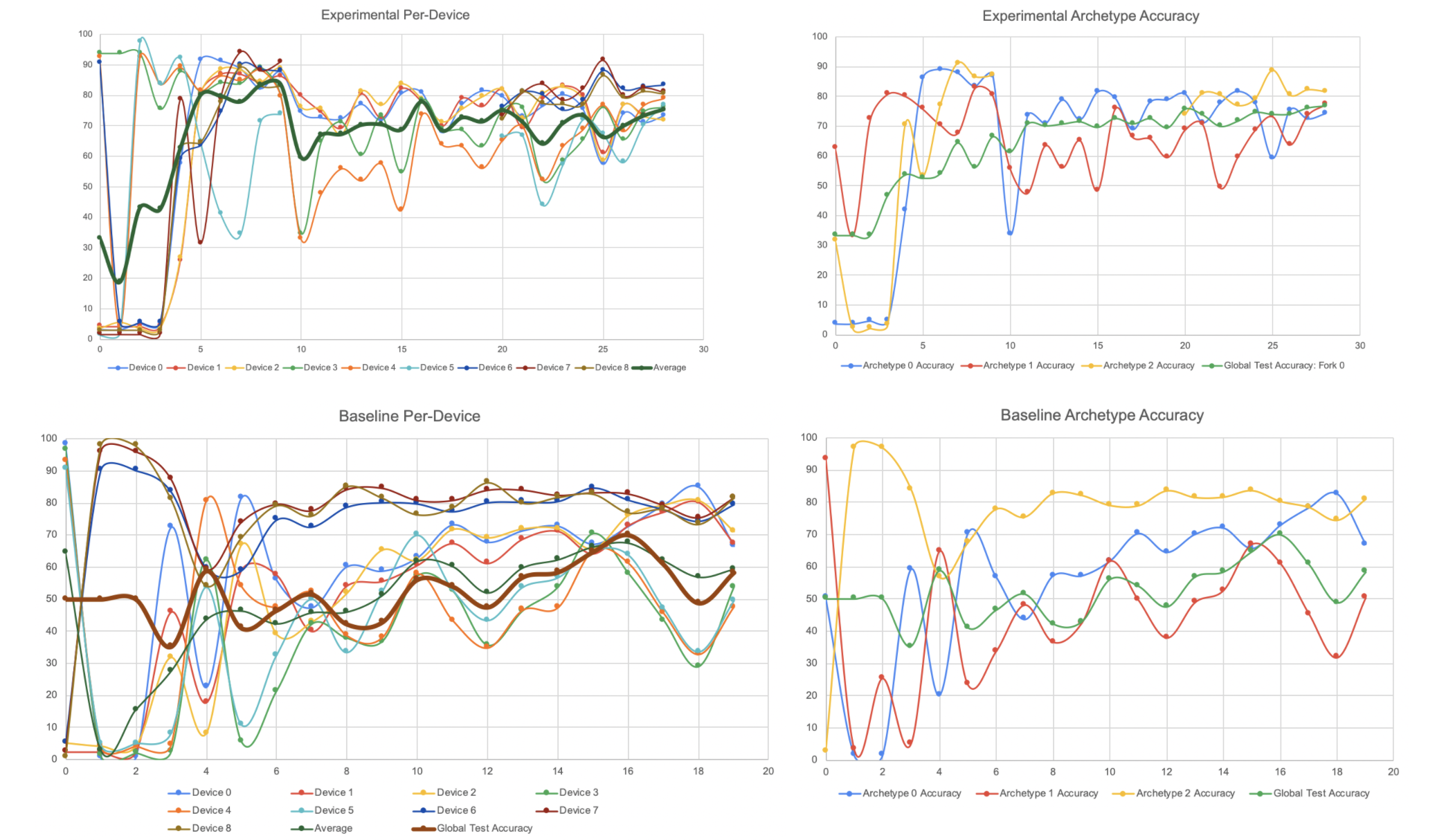}
    \caption{FedFMC versus FedAvg tested on three archetypes}
    \label{fig:ewc-results}
\end{figure}

Figure \ref{fig:ewc-results} depicts the results of the FedFMC algorithm on the simple 3-archetype dataset as compared to the Federated Averaging baseline. It's clear that FedAvg has relatively converged compared to the baseline, which has significant oscillations. Between both the forking stages and the merging stages, the performance of the different devices and archetypes are more in-line with each other rather than jumping significantly from round-to-round. 

\subsection{Grouped Archetypes: 10 Labels}
In this setup, we simulated devices that had three archetypes, with data drawn iid from the labels [0,1,2,3], [4,5,6], and [7,8,9]. An example of a real-world situation where this type of data may be present is in regional hospitals in different parts of the country, where the composition of patients is relatively uniform within a subgroup of the overall population. 

\begin{figure}[H]
    \centering
    \includegraphics[width=0.7\textwidth]{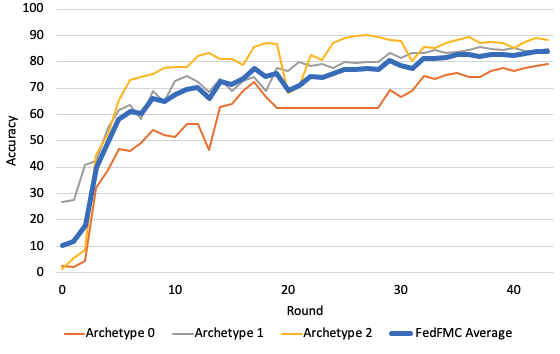}
    \caption{Performance of FedFMC on each archetype}
    \label{fig:grouped}
\end{figure}

\begin{figure}[H]
    \centering
    \includegraphics[width=0.7\textwidth]{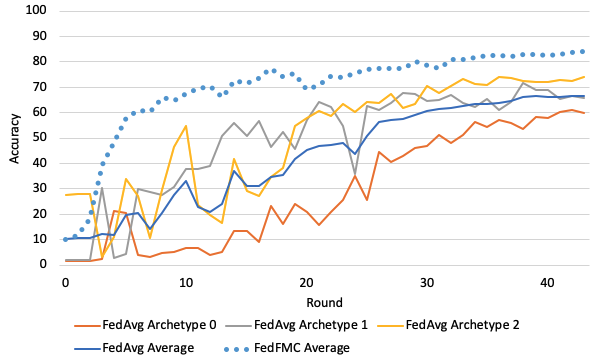}
    \caption{Performance of FedAVG on each archetype as compared to the FedFMC average}
    \label{fig:grouped2}
\end{figure}

As we can see in Figures \ref{fig:grouped} and \ref{fig:grouped2}, the performance of the FedFMC algorithm was consistently better than the performance of the FedAvg algorithm during the Forking stage (from rounds 0-24) and the merging stage, where the different global group models were consolidated into a single model. 

\subsection{Communication Costs}

The forking rounds has similar communication costs to the FedAvg algorithm with the overhead of determining the best model for devices that are performing poorly under their current group's model. The number of updates is as follows:
\begin{equation}
    u = \text{FedAvg Updates}\cdot \text{Number of Active Devices} \cdot \text{Number of Forking rounds} = E*K*T
\end{equation}
The magnitude of communication (defined by the number of times a model's weights are sent from server to local device) during the Forking phase is as follows:

\begin{equation} \label{eq3}
\begin{split}
    c &= \text{FedAvg Communication} + \text{Extra Communication to Move Models}\\ 
    &= T\cdot(2K+N)+\sum_{\text{\# forking-eligible rounds}}^{}\mathbb{E}[\text{\#devices to move}]\cdot (\text{\# active groups - 1})\\
    &\leq T\cdot (2K+N) + \sum_{t=1}^{\lfloor T/4\rfloor} N\cdot (t-1) 
\end{split}
\end{equation}

We see that in the worst-case scenario, the communication costs are $O(T^2\cdot N)$ as compared to the FedAvg communication costs of $O(TK+TN)$. In practice, even though the opportunity is presented to fork many times, the devices have been sorted into correct or approximately-correect groups such that a new group is not added at every opportunity. 

\subsection{Model Size}

One potential concern over solutions for learning non-iid data is an increase in model size. Other solutions have also utilized multiple models on each device, which linearly increases the amount of memory demanded on devices. This is a problem as many edge devices have limited memory storage.

However, we note that FedFMC does not  increase the model size. Through the {\sc Fork} and {\sc Merge} algorithms, each device only stores one model at a time (although that model does change).  

\section{Conclusion}
Federated learning has shown promise as an approach in recent years to address learning problems with privacy constraints. One of the biggest barriers to adoption is the poor experimental performance of traditional algorithms like FedAvg in situations without full device participation and without peer-to-peer edge device connections. Our algorithm, FedFMC, forks edge devices that perform similarly to update different global models, which are then consolidated using EWC through a merging phase. 
\subsection{Discussion}
There were many tuneable parameters of FedFMC algorithm, including, but not limited to:
\begin{itemize}
    \item The criteria for forking eligibility: currently the criteria is defined as at least 5 rounds from the start, more than 5 rounds from the end of forking, and at least 4 rounds between forking milestones. 
    \item The threshold for an edge device moved from its original group: currently, if $l_n^{(t)}-(\text{\upshape min }l_i^{(t)}\text{\upshape of all devices }i \in g_n^{t})> h_f*\sigma(l_i^{(t)})$ then the model is moved to the group with the best-performing model on its local dataset (and if that group is the same as the original group, it gets moved into a new group). 
    \item The criteria at which to stop the merging of a certain group: currently, a group stopped merging into the global model if the average accuracy of the global model over the previous 5 rounds was less than 1 away from the maximum accuracy of the global model over the previous five rounds of merging. 
\end{itemize}

The optimal functions and conditions for these parameters were found through experimentation. 

\subsection{Our Contribution}
The goal of this new approach is to develop a new FL scheme that can adequately learn a single model that performs well on all archetypes without significant communication costs. 

By terminating the FedFMC algorithm at the end of the Forking stage, one would get personalized models that work well for a particular archetype while still learning from all the data and not explicitly defining which devices fall into which archetype. 

By allowing the FedFMC algorithm to continue to completion with the Merging stage, one would get a single model that performs well on all of the individual local archetypes. 

\subsection{Future Work}
We also forsee several extensions of our work. Our proposed FedFMC algorithm breaks through current barriers in both Federated Learning and Lifelong Learning by through initial proof of concept experiments. The dynamic nature of FedFMC presents difficulties in using classical analysis but we would like to see further work in formalizing the intuition of forking and consolidation. In particular, developing a rigorous approach to bound the number of rounds needed for {\sc Fork} and convergence in {\sc Merge-Consolidate} proves to be nontrivial yet useful.

Besides from analysis, there are many promising areas of improvement for the FMC algorithm. On top of finetuning parameters, there is great potential in experimentation with smarter {\sc Fork} algorithms and optimizing the {\sc Merge-Consolidate} process. Experimentation is needed to find forking algorithms that quicken separation devices of different archetypes -- there are a wide variety of forking algorithms to choose form, ranging from heuristic to learned models. 

Other avenues of research extending from FedFMC include addressing open FL questions in fairness and robustness. Through the dynamic forking process, FedFMC may hold advantages over existing approaches in ensuring fair device representation. Moreover, {\sc Fork} presents devices a mechanism to self-distance from malicious devices in the network. This empowers robustness in defense against data poisoning and Byzantine attacks.

\bibliographystyle{unsrt}  
\bibliography{references}  

\end{document}